\pdfoutput=1

\documentclass[11pt]{article}

\usepackage[final]{acl}
\usepackage{times}
\usepackage{latexsym}
\usepackage{arabtex}
\usepackage{utf8}
\usepackage{todonotes}
\usepackage[shortlabels]{enumitem}

\usepackage[T1]{fontenc}

\usepackage[utf8]{inputenc}

\usepackage{microtype}

\usepackage{inconsolata}

\usepackage{graphicx}
\usepackage{multirow}
\title{Performance Analysis of Speech Encoders for Low-Resource SLU and ASR in Tunisian Dialect}
 \author{Salima Mdhaffar\textsuperscript{1}, Haroun Elleuch\textsuperscript{1,2}, Fethi Bougares\textsuperscript{2}, Yannick Estève\textsuperscript{1} \\
         \\ \textsuperscript{1}LIA, Avignon University, France\\ \textsuperscript{2}Elyadata, Paris, France  \\  salima.mdhaffar@univ-avignon.fr \\} 

\begin{document}
\maketitle

    \begin{center}
        \textbf{Abstract}
    \end{center} 
    Speech encoders pretrained through self-supervised learning (SSL) have demonstrated remarkable performance in various downstream tasks, including Spoken Language Understanding (SLU) and Automatic Speech Recognition (ASR). For instance, fine-tuning SSL models for such tasks has shown significant potential, leading to improvements in the SOTA performance across challenging datasets.
    In contrast to existing research, this paper contributes by comparing the effectiveness of SSL approaches in the context 
    of (i) the low-resource spoken Tunisian Arabic dialect and (ii) its combination with a low-resource SLU and ASR scenario, where only a few semantic annotations are available for fine-tuning. 
    We conduct experiments using many SSL speech encoders on the TARIC-SLU dataset. We use speech encoders that were pre-trained on either monolingual or multilingual speech data. Some of them have also been refined without in-domain nor Tunisian data through multimodal supervised teacher-student paradigm.
    This study yields numerous significant findings that we are discussing in this paper. 

\section{Introduction}
\label{sec:intro}
Self-supervised learning methods aim to train a representational model, also called upstream model, that benefits a collection of downstream tasks. 
SSL models are trained by using information extracted from the input data itself as the label to target. 
Various techniques have been introduced in the literature in order to learn powerful representations from the speech signal, including those based on autoregressive predictive coding \cite{chung2019unsupervised}, contrastive losses \cite{schneider2019wav2vec,baevski2020wav2vec} and masked prediction \cite{liu2021tera,chen2022wavlm,hsu2021hubert}.
Other works explored the combination of contrastive learning and masked language modeling \cite{chung2021w2v}.
The learned SSL models like wav2vec 2.0 \cite{baevski2020wav2vec}, wavLM \cite{chen2022wavlm}, data2vec \cite{baevski2022data2vec}, data2vec 2.0 \cite{Baevski2022EfficientSL}, ArTST \cite{toyin2023artst}, w2v-BERT \cite{chung2021w2v} have been proven effective for a wide range of tasks as they considerably lighten the amount of annotated speech data normally required for downstream tasks, while exploiting very large amounts of unlabeled data. Some models  have been extended to a cross-lingual setting through XLS-R-128 \cite{babu2021xls}, MMS \cite{pratap2023scaling} and the recently released w2v-BERT 2.0 model \cite{barrault2023seamless}. Several efforts went further to enrich the frame-level speech representations with textual semantic information.
In 2022, \citet{khurana2022samu} proposed such a model called SAMU-XLSR allowing the obtention of a better semantic encoding in the speech representations of a pre-trained SSL model. 
SAMU-XLSR approach follows a teacher-student supervised learning framework that uses multilingual text/audio paired data and the Language-agnostic BERT Sentence Embedding (LaBSE) model~\cite{feng2022language}. 
This teacher-student approach is used to refine an SSL pre-trained model dedicated to speech processing to make it able to produce sentence-level embeddings from speech similar to the embbedings given by the LaBSE model from the speech transcriptions.
They use the CommonVoice version 8 corpora applied to 53 languages to refine the XLS-R-128 model~\cite{babu2021xls}. 

More Recently, a multilingual sentence embedding space for 200 text and 37 speech languages called SONAR \cite{duquenne2023sentence} was introduced by META AI. The authors adopted a two-step training strategy: They first build a sentence embedding space for the multilingual text before extending this to the speech modality via a teacher-student approach.
Applied to many downstream tasks, SSL speech encoders have shown their great potential by improving the state-of-the-art performances on challenging benchmark datasets. 
 However, understanding speech encoder capabilities requires a benchmarking effort  to compare and draw insights across
the techniques.
Indeed, there has been a considerable amount of work and effort in order to benchmark such SSL models. 
SUPERB \cite{yang2021superb} is a good example of this effort. 
It provides a comprehensive speech SSL benchmark including tasks such as phoneme detection, ASR, slot filling, intent detection, keyword spotting, etc. 
The SUPERB benchmark has been extended to evaluate multilingual speech systems in the ML-SUPERB benchmark \cite{shi2023ml}. 
XTREME-S \cite{conneau2022xtreme} is another example that focuses only on SSL models trained with multiple languages. 
Despite the large number of high-quality benchmarks that evaluate SSL models on various downstream tasks, a limited number of studies have probed their effectiveness for downstream tasks in spoken Arabic dialects. 

In this paper, we propose to broaden the SSL benchmarking effort to the Automatic Speech Recognition (ASR) and Spoken Language Understanding (SLU) of spoken Arabic Dialects. The contributions of this work are as follows: 
\begin{enumerate} 
    \item We benchmark the effectiveness of various state-of-the-art SSL speech encoders in the very challenging context of a low-resource spoken Arabic dialect (Tunisian dialect) with a limited training data;
    \item We compare the performances of mono vs. multilingual/bilingual SSL models, and the impact of a semantic encoding refinement through a multimodal supervised teacher-student approach;
    \item We explore the use of recently released SSL models (w2v-BERT 2.0 and SONAR) for both ASR and SLU. To our knowledge, this is the first work in the literature to evaluate w2v-BERT 2.0 and SONAR for an SLU task;
    \item We release our code and models\footnote{\url{https://github.com/speechbrain/speechbrain/tree/develop/recipes/TARIC}} for reproducibility and to encourage future research on ASR and SLU of Arabic dialects.
\end{enumerate}

\section{Benchmarking protocol}
\label{sec:pagestyle}

In this section, we formulate the downstream tasks and  describe the Tunisian dialect dataset together with the speech encoders used in our study.

\subsection{Downstream tasks}
In this work, we address two tasks: Automatic Speech Recognition and Spoken Language Understanding  of the Tunisian Arabic spoken dialect. \\

\begin{enumerate}[(a)]
    \item ASR: Automatic Speech Recognition has the ultimate goal of providing the correct transcription given spoken utterance. It is used to assess the ability of SSL models to extract content information from audio inputs. The ASR task will be evaluated in terms of Word Error Rate (WER).
    \item SLU: Spoken language understanding refers to natural language processing tasks that aim to extract semantic information from speech \cite{tur2011spoken}.
Different tasks can be addressed as SLU tasks, such as named entity recognition from speech, dialog state tracking, intent recognition, slot filling, etc\ldots
In the context of a conventional dialogue system, information is typically represented through a semantic frame structure. 
For each utterance, constructing the semantic representation primarily involves (i) classifying the user's utterance in terms of `{speech acts}’ (SA)~\cite{searle_1969} or `{intents}’ and (ii) slot filling~\cite{wang2005spoken}.
We consider these two SLU tasks following the available annotations: (1) \textbf{Speech Act classification} classifies speech utterances into predefined classes to determine the intent of the speaker; (2) \textbf{Slot filling} is a natural language processing and information extraction technique that entails the identification and extraction of specific pieces of information or attributes, referred to as `slots', from unstructured text or spoken language. These slots are typically associated with predefined categories or entities.
The SLU speech act recognition will be evaluated in terms of Speech Act Error Rate (SAER), which is a standard classification error rate.
The SLU slot filling task will be evaluated in terms of Concept Error Rate (COER) and Concept/Value Error Rate (CVER).
COER is computed similarly to WER by taking into account only the semantic labels in the reference and hypothesis annotations. 
The CVER computation is identical, but the occurrences of concept/value pairs are taken into account instead of the concept alone. 
CVER implies that if any character within the word's support prediction or the concept tag prediction differs from the reference, the entire prediction for that concept is considered as an error. 
For the calculation of these metrics, we have drawn on the detailed description in~\citet{laperriere2022spoken}.
\end{enumerate}


\subsection{Dataset}


The ASR dataset TARIC \cite{masmoudi2014phonetic} has been used for this work along with its recent SLU enrichment, TARIC-SLU~\cite{mdhaffar2024taric-slu}.
The acquisition of the TARIC dataset was carried out in train stations in Tunisia. 
The dataset is made of human-human recordings with their manual transcriptions and semantic annotations.
It is composed of more than 2,000 dialogues from 109 different speakers.
The dataset\footnote{\url{https://demo-lia.univ-avignon.fr/taric-dataset/}} is split into three parts (train, dev and test) as described in Table \ref{tab:taric_data_split}.

\begin{table}[ht]
\centering
\caption{TARIC-SLU data set split into Train, Dev and Test}

\begin{tabular}{llll}
\hline 
            & Train & Dev & Test \\ \hline \hline
\#Utterance & 15752   & 771    & 1249 \\
\#Dialog    & 1713    & 103    &  173  \\ 
Duration    & 7.5 hrs & 29 min & 53 min   \\ \hline
\end{tabular}
\label{tab:taric_data_split}
\end{table}

TARIC-SLU was annotated using 62 semantic concept tags such as \textit{city name arrival}, \textit{departure time}, \textit{ticket price}, etc. and 3 speech acts (\textit{directives-answer}, \textit{directives-query} and \textit{politeness}). 
The example below shows the original sentence (a) together with its English translation (b) and the corresponding semantic annotation (c).  

As for the annotation tags, the first word of each sentence represents the speech act tag (\speechact{color_concept_a}{Directives-query}{} in this example). 
`\openconcept{color_concept_e}{city-name-departure}' is an opening tag starting the support word sequence `Sfax' and expressing that this word sequence is associated with the \textsl{city-name-departure} semantic concept. The character \textit{`$>$'} represents the closing tag, and it is used to close all concept tags. 
The same schema is used for the other semantic concepts. 
For speech act tags, there is no closing tag, since each sentence has a single speech act.

\begin{enumerate}[(a)]
\item  \setcode{utf8} \begin{arabtext} \RL{باللاهي أعطيني زوز تكايات من صفاقس لتونس مع الثمنية و نصف} \end{arabtext}
\item Please give me two tickets from Sfax to Tunis at eight thirty
\item \speechact{color_concept_a}{Directives-query}{} please
        \concept{color_concept_b}{command-task}{give me} \concept{color_concept_c}{number-of-tickets}{two}  \concept{color_concept_d}{object}{tickets} from
        \concept{color_concept_e}{city-name-departure}{Sfax} to \concept{color_concept_f}{city-name-arrival}{Tunis} at 
        \concept{color_concept_g}{departure-time}{eight thirty}
\end{enumerate}

\subsection{SSL speech encoders}

\begin{table*}[h!]
\centering
\caption{Architecture details of benchmarked SSL encoders. *~SAMU-XLSR is a modified version of XLS-R-128,  $^\clubsuit$ all SONAR speech encoders models are a modified version of w2v-BERT, $^\star$ represents the size of the paired dataset used in the refinement step.}
\begin{tabular}{|l|l|l|l|l|}
\hline
Speech encoder    & \#Param & \#Lang &  \ Hours &  Network architecture\\ \hline \hline

LeBenchmark-7K \cite{evain2021task}  & 317M & 1    &   7K   & 7CNN-24Trans-Enc \\ 
English lv60  \cite{baevski2020wav2vec} &  317M& 1 &  960  & 7CNN-24Trans-Enc \\ 
HuBERT \cite{hsu2021hubert}             & 316M & 1    &  60K & 7CNN-24Trans-Enc \\ 
wavLM  \cite{chen2022wavlm}             & 316M & 1    &  94K  & 7CNN-24Trans-Enc \\ 
data2vec 2.0 \cite{Baevski2022EfficientSL}     & 314M    & 1    &  960  & 7CNN-24Trans-Enc\\
VP-100K \cite{wang2021voxpopuli}& 317M &23&  100K  & 7CNN-24Trans-Enc \\ 
XLS-R-128 \cite{babu2021xls}& 317M & 128  &   436K & 7CNN-24Trans-Enc \\ 
MMS\cite{pratap2023scaling}          & 317M & 1024 &  23K  & 7CNN-24Trans-Enc \\ 
MMS-1B \cite{pratap2023scaling}         & 1B   & 1024 &  23K & 7CNN-48Trans-Enc\\ 
w2v-BERT 2.0 \cite{barrault2023seamless} & 600M & 143 & 4.5M  &  2CNN-24Conformer\\ \hline \hline
SAMU-XLSR \cite{khurana2022samu} $^*$   & 317M & 128  &  436K+12.7K$^\star$ & 7CNN-24Trans-Enc \\ 
SONAR-ARB \cite{duquenne2023sentence} $^\clubsuit$  & 600M & 2 & 60K+822$^\star$ & 2CNN-24Conformer\\
SONAR-ENG \cite{duquenne2023sentence} $^\clubsuit$  & 600M & 1 & 60K+N/A$^\star$ & 2CNN-24Conformer\\
SONAR-FRA \cite{duquenne2023sentence} $^\clubsuit$  & 600M & 2 & 60K+2K$^\star$ & 2CNN-24Conformer\\
\hline
\end{tabular}
\label{tab:ssl_models}
\end{table*}

\begin{table*}[h!]
\centering
\caption{Architecture details of Whisper models}
\begin{tabular}{|l|l|l|l|l|}
\hline
Model & Network architecture & \#Params \\ \hline 
Whisper-small  \cite{radford2023robust}    & 2-conv 12 Enc-Dec & 244 M \\ 
Whisper-medium \cite{radford2023robust} & 2-conv 24-Enc-Dec & 769 M \\ \hline
\end{tabular}
\label{tab:whispermodels}
\end{table*}

For our study, we used different types of speech encoders.
We considered monolingual SSL speech encoders (French: wav2vec 2.0 LeBenchmark-7K; English: wav2vec 2.0 LV60, HuBERT, wavLM, data2vec  2.0 and SONAR-ENG) as well as cross-lingual SSL speech encoders (wav2vec 2.0 VP-100K, XLS-R-128, MMS, MMS-1B, w2v-BERT 2.0 and SAMU-XLSR) and two bilingual speech encoders (SONAR-ARB and SONAR-FRA). 
As explained in the introduction, SAMU-XLSR is a modified version of XLS-R-128 and SONAR models are a modified version of w2v-BERT \footnote{Notice that w2v-BERT and w2v-BERT 2.0 are two different models. Both of them are trained using the same architecture, but w2v-BERT is an English model and w2v-BERT 2.0 is a multilingual model. w2v-BERT has not been released to the research community, so we cannot evaluate its performance.}.
We put the URLs for all these models in the Appendix \ref{sec:appendix}.

\begin{figure}[h!]
\centering
\includegraphics[scale=0.45]{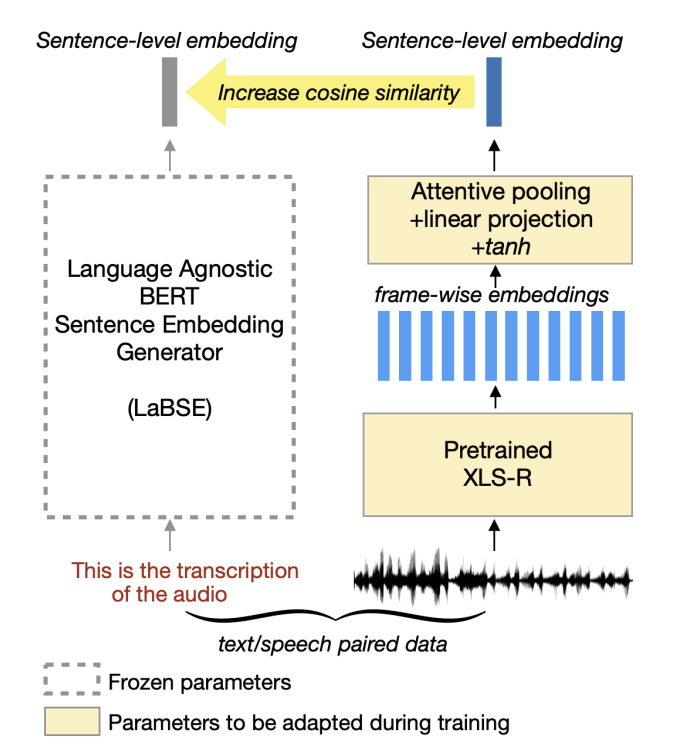}
\caption{SAMU-XLSR Training and specialization} 
\label{fig:samu-xlsr}
\end{figure}

SAMU-XLSR is based on the pre-trained multilingual XLS-R-128. SAMU-XLSR will process  audio and text paired data. The  XLS-R-128 model used in this approach was designed to generate speech representations for short 20 milliseconds speech frames. To make use of this model, SAMU-XLSR performs pooling and projection to create a single sentence-level representation. 
In parallel, LaBSE \cite{feng2022language} sentence-level textual representations are simply extracted. Both representations being on the same semantic space, SAMU- XLSR’s is then being pulled towards LaBSE’s with the help of a cosine similarity loss function. This means the parameters of all SAMU-XLSR’s components are optimized to predict the textual representations generated by the frozen LaBSE model.
Figure \ref{fig:samu-xlsr} illustrates the training process of SAMU-XLSR.

\begin{figure}[h!]
\centering
\includegraphics[scale=0.25]{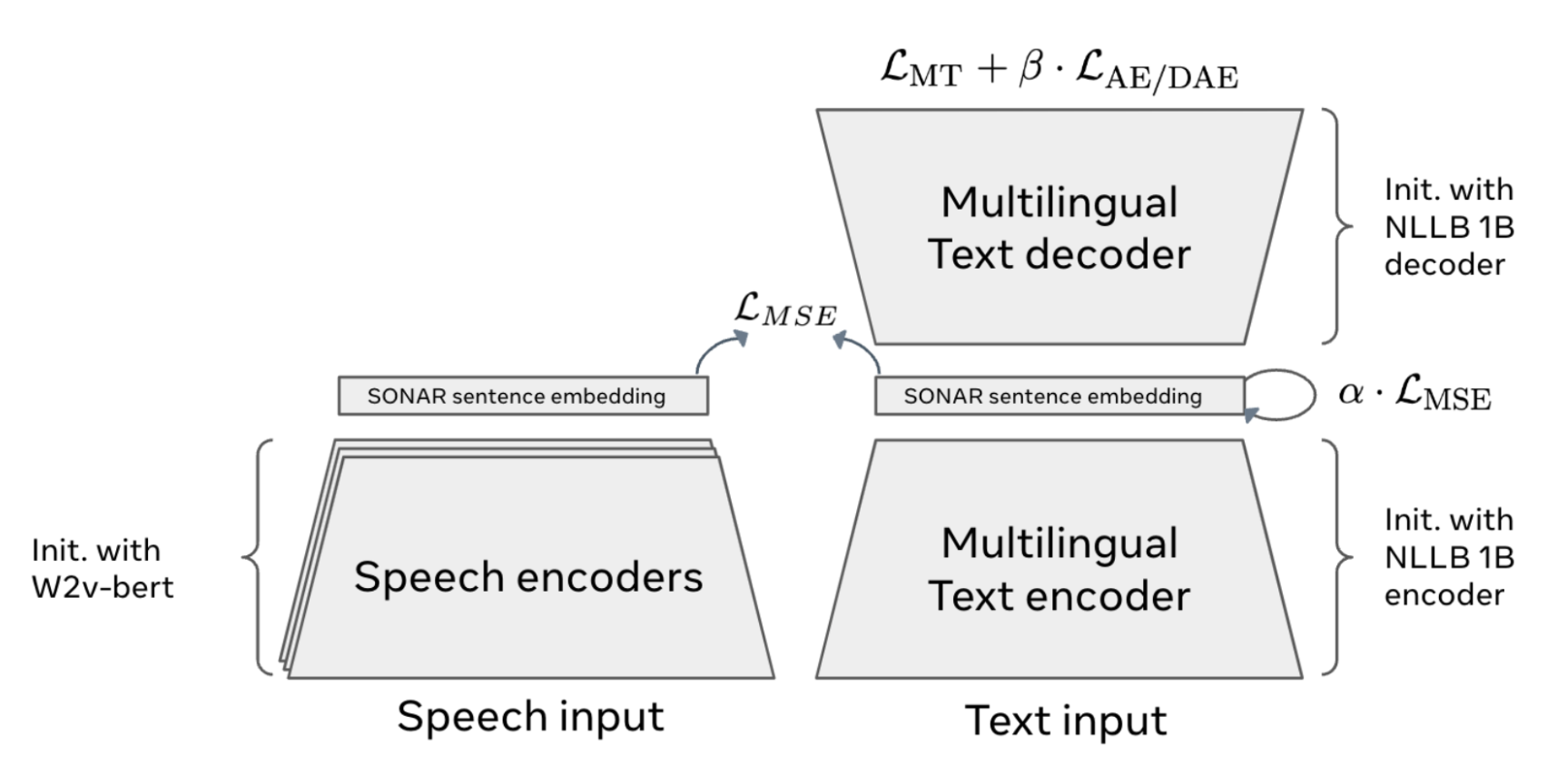}
\caption{Sonar architecture} 
\label{fig:sonar}
\end{figure}

SONAR is a new  multilingual and -modal text embedding space trained in encoder-decoder architecture for 200 languages, which substantially outperforms existing approaches like LASER3 \cite{heffernan2022bitext} or LaBSE in multilingual similarity search. Authors apply a teacher-student approach to extend this embedding space to the speech modality and currently cover 36 languages. Mining is performed in data from publicly available repositories of web data (tens of billions of sentences) and speech (4 million hours). In total, to train the model, they align more than 443,000 hours of speech with texts and create about 29,000 hours of speech-to-speech alignments. Figure \ref{fig:sonar} illustrates the training process of SONAR.

Overall, we experimented with 14 different speech encoders.
Table \ref{tab:ssl_models} shows all the details of SSL speech encoder models. \\
In addition to the aforementioned SSL models,
we also evaluated using the recently released Whisper models from OpenAI \cite{radford2023robust}.
Unlike self-supervised speech models, Whisper is a multilingual ASR model trained using a large amount of labelled audio transcription data (680k hours).
Using Whisper was motivated by the recent achievements reported by \citet{wang2023whislu}.
The detailed properties of Whisper models used in this study are presented in Table \ref{tab:whispermodels}.

\begin{table*}[ht]
\centering
\caption{Comparison of speech encoders performance across ASR and SLU tasks. Results are reported in WER for ASR and SLU system transcripts. SLU results are reported using COER and CVER for slot filling detection and in terms of SAER for speech act classification. Bold numbers are the best results in all the table. Underlined numbers represent the best results for each colored block (mono-lingual models, cross-lingual models without teacher-student multi-modal training, cross-lingual models with teacher-student multi-modal training, whisper models). FR-7K refers to LeBenchmark-7K, Whisper-S refers to Whisper-Small, Whisper-M refers to Whisper-Medium} 
\begin{tabular}{|l|ll|ll|ll|ll||ll|}
\hline
\multirow{1}{*}{Speech encoder} & \multicolumn{2}{l|}{COER (SLU) }       & \multicolumn{2}{l|}{WER (SLU) }        & \multicolumn{2}{l|}{CVER (SLU) }      & \multicolumn{2}{l||}{SAER (SLU) }  &  \multicolumn{2}{l|}{WER (ASR) } \\ \cline{2-11}     
                    & \multicolumn{1}{l|}{Dev} & Test & \multicolumn{1}{l|}{Dev} & Test & \multicolumn{1}{l|}{Dev} & Test  & \multicolumn{1}{l|}{Dev} & Test  & \multicolumn{1}{l|}{Dev} & Test \\ \hline \hline 

\rowcolor{LightCyan}
FR-7K                & \multicolumn{1}{l|}{37.78}    &  33.16    & \multicolumn{1}{l|}{36.04}    & 28.61     & \multicolumn{1}{l|}{57.57}    &   49.57 & \multicolumn{1}{l|}{\underline{\textbf{23.87}}}    &   \underline{21.14}  &  \multicolumn{1}{l|}{34.80}  &  29.69 \\ 
\rowcolor{LightCyan}
English lv60                           & \multicolumn{1}{l|}{36.77}    &  32.69    & \multicolumn{1}{l|}{37.39}    &  30.49    & \multicolumn{1}{l|}{57.7}    &  50.04  & \multicolumn{1}{l|}{26.07}    &    21.94 &  \multicolumn{1}{l|}{35.04}  & 30.2  \\ 
\rowcolor{LightCyan}
HuBERT                             & \multicolumn{1}{l|}{39.84}    &   33.76   & \multicolumn{1}{l|}{39.41}    &  31.74    & \multicolumn{1}{l|}{61.16}    &  52.80  & \multicolumn{1}{l|}{25.94}    &   21.86  &  \multicolumn{1}{l|}{34.25}  &  28.75 \\ 
\rowcolor{LightCyan}
WavLM                             & \multicolumn{1}{l|}{35.85}    &  32.25    & \multicolumn{1}{l|}{\underline{34.22}}    &  \underline{27.23}    & \multicolumn{1}{l|}{\underline{55.95}}    &   \underline{50.92} & \multicolumn{1}{l|}{24.9}    &   \underline{21.14}  &  \multicolumn{1}{l|}{32.28}  & \underline{26.7} \\
\rowcolor{LightCyan}
Data2vec 2.0                            & \multicolumn{1}{l|}{\underline{34.50}}    &  \underline{31.8}    & \multicolumn{1}{l|}{39.41}    &   31.71   & \multicolumn{1}{l|}{58.94}    &  51.15  & \multicolumn{1}{l|}{25.16}    &  22.9  &  \multicolumn{1}{l|}{38.58}  & 32.1  \\ 
\rowcolor{pink}
VP-100K                  & \multicolumn{1}{l|}{35.77}    &   31.56   & \multicolumn{1}{l|}{35.46}    &  27.56    & \multicolumn{1}{l|}{56.37}    &   48.90  & \multicolumn{1}{l|}{\underline{24.64}}    &  22.9  &  \multicolumn{1}{l|}{32.71}  & 26.68 \\
\rowcolor{pink}
XLS-R-128                  & \multicolumn{1}{l|}{35.62}  &  31.24    & \multicolumn{1}{l|}{34.65}    &   26.7   & \multicolumn{1}{l|}{56.24}    &   48.73 & \multicolumn{1}{l|}{\underline{24.64}}    &   \underline{\textbf{20.9}}  &  \multicolumn{1}{l|}{33.82}  &  28.06 \\ 
\rowcolor{pink}
MMS                                & \multicolumn{1}{l|}{36.73}    &   31.77   & \multicolumn{1}{l|}{43.97}    &  37.98    & \multicolumn{1}{l|}{62.07}    &  56.47 & \multicolumn{1}{l|}{24.9}    &   24.66  &  \multicolumn{1}{l|}{35.91}  & 29.46  \\
\rowcolor{pink}
MMS-1B                             & \multicolumn{1}{l|}{43.82}    &   36.13   & \multicolumn{1}{l|}{44.36}    &  38.46    & \multicolumn{1}{l|}{66.91}    &  58.03  & \multicolumn{1}{l|}{28.4}    &   23.83  &  \multicolumn{1}{l|}{41.97}  & 34.76 \\
\rowcolor{pink}
w2v-BERT 2.0                          & \multicolumn{1}{l|}{\underline{\textbf{32.29}}}    &  \underline{\textbf{29.13}}  & \multicolumn{1}{l|}{\underline{\textbf{26.08}}}    &  \underline{\textbf{20.84}}    & \multicolumn{1}{l|}{\underline{\textbf{49.55}}}    & \underline{\textbf{46.22}}  & \multicolumn{1}{l|}{25.6}    &  \underline{\textbf{20.9}} & \multicolumn{1}{l|}{\underline{\textbf{25.11}}} &  \underline{\textbf{21.47}}  \\ 
\rowcolor{yellow}
SAMU-XLSR                             & \multicolumn{1}{l|}{\underline{32.73}}    &  \underline{30.11}   & \multicolumn{1}{l|}{\underline{31.10}}    &  \underline{23.95}    & \multicolumn{1}{l|}{\underline{51.93}}    & \underline{48.06}  & \multicolumn{1}{l|}{\underline{24.12}}    &  \underline{22.5}  &  \multicolumn{1}{l|}{\underline{28.56}}  &  \underline{24.66}  \\  
\rowcolor{yellow}
SONAR-ENG                          & \multicolumn{1}{l|}{36.58}    &  33.59  & \multicolumn{1}{l|}{38.34}    &  31.58    & \multicolumn{1}{l|}{57.15}    & 52.33  & \multicolumn{1}{l|}{36.71}    &  26.10  &  \multicolumn{1}{l|}{39.77}  & 33.67 \\

\rowcolor{yellow}
SONAR-FRA                          & \multicolumn{1}{l|}{36.88}    &  33.8  & \multicolumn{1}{l|} {38.29}    &  30.38    & \multicolumn{1}{l|}{58.83}    & 53.43  & \multicolumn{1}{l|}{34.46}    &  27.79 &  \multicolumn{1}{l|}{39.16}  & 32.71  \\

\rowcolor{yellow}
SONAR-ARB                          & \multicolumn{1}{l|}{35.93}    &  31.62   & \multicolumn{1}{l|}{35.6}    &  28.17    & \multicolumn{1}{l|}{55.98}    & 49.77  & \multicolumn{1}{l|}{32.68}    &  23.38 &  \multicolumn{1}{l|}{35.24}  &  29.05 \\
 \hline \hline
 
Whisper-S                            & \multicolumn{1}{l|}{39.52}    &  34.81   & \multicolumn{1}{l|}{39.79}    &  32.85   & \multicolumn{1}{l|}{64.83}    &  56.57  & \multicolumn{1}{l|}{32.99}    &  29.56   &  \multicolumn{1}{l|}{38.5}  & 32.1\\
Whisper-M                           & \multicolumn{1}{l|}{\underline{39.1}}    &  \underline{33.96}   & \multicolumn{1}{l|}{\underline{33.37}}    &    \underline{29.56} 
& \multicolumn{1}{l|}{\underline{59.13}}    & \underline{54.02}    & \multicolumn{1}{l|}{\underline{25.94}}    &   \underline{21.86}   & \multicolumn{1}{l|}{\underline{32.5}}   & \underline{29.05} \\ \hline
\end{tabular}
\label{tab:results}
\end{table*}

\section{Experiments and Results}
\label{sec:Exp_results}

\subsection{Training details}
\label{sec:training_details}
For comparison purposes, we set the same parameters for all the models using the different speech encoders.

\subsubsection{ASR}
In addition to the speech encoder model, we incorporate an extra layer with 1024 neurons and LeakyReLU as the activation function, followed by a fully-connected layer and a final $40$-dimensional softmax layer, each dimension corresponding to a character.
The weights of these two additional layers were randomly initialized, while the weights of the speech encoder part for SSL models of the neural architecture were initialized using the pre-trained weights.
The fine-tuning is done with the TARIC training set using a character-level CTC loss function. 
We optimize the loss with an Adam optimizer of learning rate = 0.0001 for both speech encoder, and Adadelta with learning rate = 1.0 for the linear layer.

\subsubsection{SLU}
We formulate the end-to-end SLU task as a character level prediction where slots are delimited by tag-specific special characters, as in \citet{yadav2020end,Ghannay2018,mdhaffar2022impact}. 
We also added the speech act token to the reference annotation as the first token of each sequence of words. 
This way, the end-to-end model learns to both classify the utterances in terms of speech act, and recognize slot/value pairs present in the speech segment.
As input, the neural network receives a WAV audio file (PCM, 16 bits, 16kHz, signed integer), and the output is a transcription enriched with semantic labels and speech acts tags.
After processing through the softmax layer (which has the size of 106\footnote{40 characters that cover the alphabet of the TARIC dataset, 62 characters for slots, one character for closing slots and three characters for speech acts}), the outputs are generated by a simple greedy decoder.
We optimize the loss with an Adam optimizer of learning rate = 0.0001 for both speech encoder, and Adadelta with learning rate = 1.0 for the linear layer.

\subsubsection{Training details for ASR and SLU}
We employed a batch size of 4 samples, distributed across 4 NVIDIA V100 32GB GPU cards.
For MMS-1B, we used 4 NVIDIA A100 80GB GPU cards.
We utilized two optimizers: Adadelta for updating the additional layers weights and Adam for fine-tuning the self-supervised learning (SSL) model.
The initial learning rate for Adadelta was  $1.0$, while for Adam it was set to $0.0001$. 
Our models were implemented using the SpeechBrain toolkit~\cite{ravanelli2021speechbrain}.
The speech encoders are obtained through the fairseq \cite{ott2019fairseq} framework for SAMU-XLSR \footnote{SAMU-XLSR is not yet publicly available; it has been kindly shared by the authors of~\citet{khurana2022samu}}, SONAR and data2vec 2.0 and through HuggingFace for the remaining of models. 
More details about our models and the configuration files will be publicly available as a part of the Speechbrain toolkit.

\subsection{Results}
\label{sec:results}
In this section, we report, analyze and discuss the performance of our 
models across various dimensions. All the results, including both ASR and SLU evaluations, are presented in Table \ref{tab:results}. 
ASR models are evaluated with WER and SLU models are evaluated using WER (after removing the semantic tags and the speech act tag from the system output), COER, SAER and CVER. 
The first part of the table shows results by using SSL models, and the second part shows results by using Whisper models.
The first part is divided into three sub-parts according to the type of SSL models: (i) the first sub-part, colored blue for monolingual models, (ii) the second sub-part, colored pink for cross-lingual models, and (iii) the third sub-part, colored yellow for the cross-lingual model after a teacher-student multi-modal semantic training. The second part of this table is dedicated to the results obtained when using Whisper small and medium models.

Overall, except for the SLU SAER score on the dev set, the best ASR and SLU results are obtained when our models are trained using w2v-BERT 2.0 model. Below, an analysis of the obtained results according to the type of the used SSL models.

\textbf{Monolingual models:} When we compare the performances of models trained using monolingual SSL (blue section of the results table), we observe that WavLM is by far the best-performing model for the ASR task with \textbf{32.28\%} and \textbf{26.7\%} WER for dev and test sets respectively. This is confirmed as well when evaluating WER of the SLU model output (column WER (SLU) in table \ref{tab:results}). Regarding the SLU task, while we did not observe any emerged trend, we observed a better performance of data2vec 2.0 in terms of COER evaluation.

\textbf{Cross-lingual models without teacher-student multi-modal semantic training:} Results indicate that w2v-BERT 2.0 yields the best performance in both ASR and SLU tasks (expect for the SAER).

\textbf{All cross-lingual models:} When it comes to cross-lingual models, with (colored in pink) or without teacher-student multi-modal semantic training (colored in yellow), setting apart the w2v-BERT 2.0 model, SAMU-XLSR clearly outperforms VP-100k and MMS based models.

\textbf{Mono-lingual vs. cross-lingual models:}
If we compare performances between monolingual and cross-lingual SSL models, cross-lingual models achieve better results for the SLU task. Indeed, data2vec 2.0 shows competitive COER scores while its WER is higher compared to other models. This leads to the conclusion that while data2vec 2.0 demonstrates the ability to identify semantic concepts, it falls short in providing accurate transcriptions.

\textbf{Cross-lingual models with teacher-student refinement: }
As stated earlier, the SLU performances of VP-100K, XLSR-128, and MMS exhibit a consistent trend, with XLSR-128 showing the highest performance among them. 
Combining this multilingual SSL (XLS-R-128) and teacher-student multimodality training in one model (SAMU-XLSR) provides better results for both ASR and SLU.
Shifting the focus to SONAR models, as expected, SONAR-ARB demonstrates superior performance compared to both SONAR-ENG and SONAR-FRA for both ASR and SLU tasks. SONAR models are based on the English version of w2v-BERT. Unfortunately, the w2v-BERT model used to initialize the speech encoders of SONAR is not available to assess its contribution.

\textbf{Whisper models:} Results show that Whisper-medium outperform results obtained using Whisper-small. 

\textbf{Whisper models vs. all SSL models:} If we compare performances between Whisper models (second part of the table) and all SSL models (first part of the table), almost SSL models achieve better results for the slot filling SLU task. To transcribe audio files, WER obtained by the ASR or the SLU show competitive results. For example, ASR results for the dev set (32.5\%) obtained by Whisper medium outperform almost all SSL models except for w2v-BERT 2.0, SAMU-XLSR and wavLM.
This leads to the conclusion that while Whisper models demonstrates the ability to transcribe audio files, it falls short in providing accurate semantic extraction compared to the use of SSL speech encoders.

\section{Discussion}
We carried out some error analyses to quantify some SLU systems performance across different conditions.
We focus on 6 SSL models: LeBenchmark, XLS-R-128, data2vec 2.0, SAMU-XLSR, w2v-BERT 2.0 and SONAR-ARB. 

\subsection{Acoustic complexity}

\begin{figure}[h!]
\centering
\includegraphics[scale=0.2]{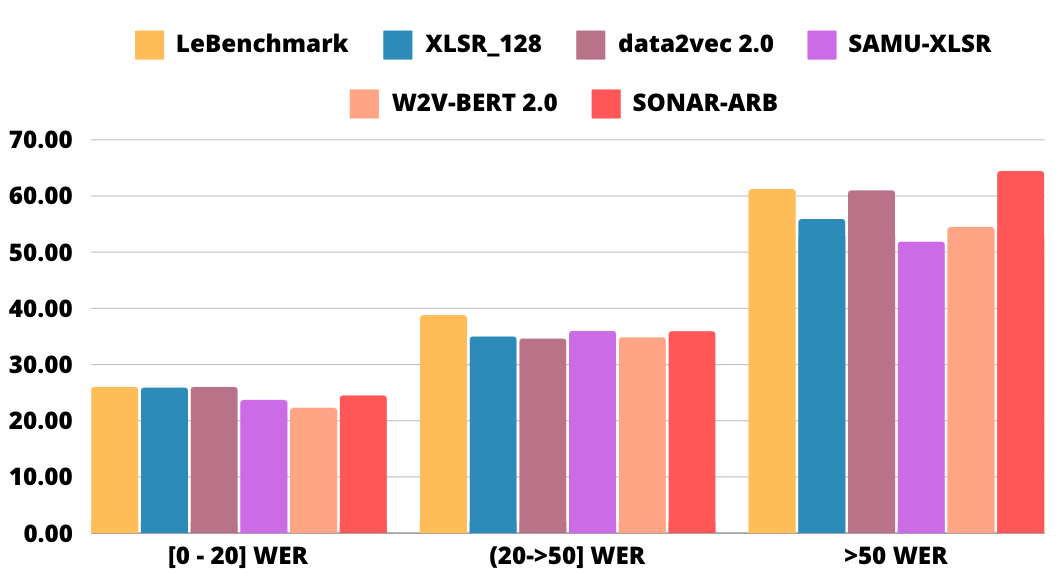}
\caption{SLU performance (COER) across different general complexity levels in test utterances.} 
\label{fig:slots}
\end{figure}

First, we used the WER of the transcripts produced by the model giving the best ASR system (w2v-BERT 2.0) to quantify the general complexity of the utterance. 
We defined three groups of segments with low (WER $<=$ 20), medium (20 $<$ WER $<=$ 50)  and high (WER $>$ 50)  complexity.
Figure~\ref{fig:slots} shows that, as expected, the SLU performances are overall better for spoken utterances belonging to the low complexity group. 
The x-axis represents different levels of WER, and the y-axis represents COER.
However, when we compare the COER of the SLU systems by group, we observe that w2v-BERT performs better for segments with low WER. For segments with medium WER, we observe that all the systems have a comparable behavior except for LeBenchmark: the system shows a higher COER. For utterances hard to transcribe, SAMU-XLSR is the best one followed by w2v-BERT 2.0.

\subsection{Semantic complexity}

\begin{figure}[h!]
\centering
\includegraphics[scale=0.2]{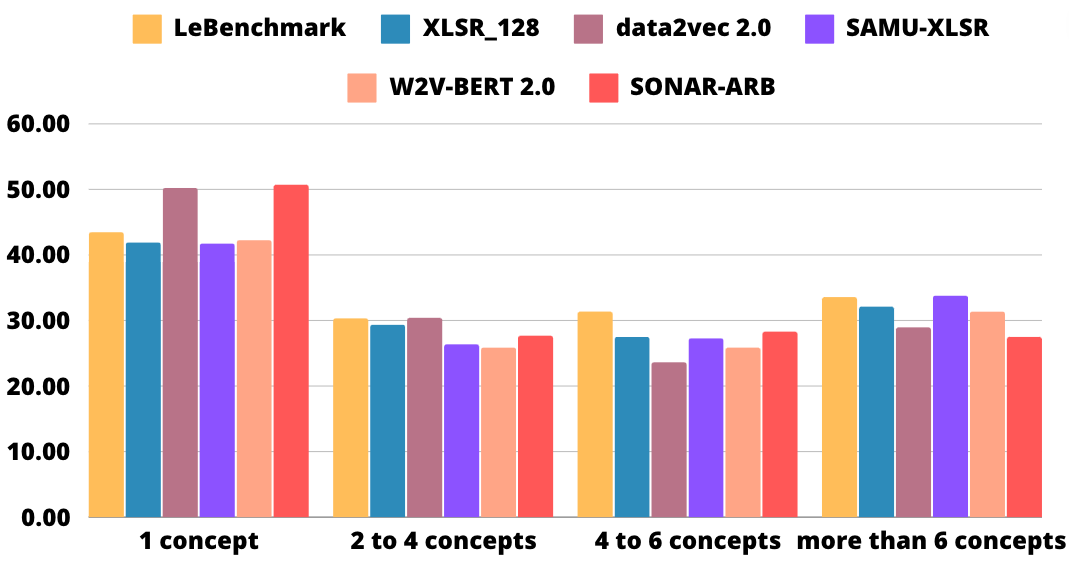}
\caption{SLU performance (COER) across different numbers of semantic tags in test utterances.} 
\label{fig:concepts}
\end{figure}

In the semantic complexity analysis, we used the number of semantic tags per utterance in the reference as a proxy of the semantic complexity. 
Figure \ref{fig:concepts} shows the results obtained by the evaluated models. 
The x-axis represents the number of semantic tags in test utterances, and the y-axis represents COER.
Across the board, utterances with two to six concepts seem to be the easiest.  
While data2vec 2.0 and SONAR-ARB perform worse when there are fewer semantic concepts to extract. They perform best when there are more than six.


\section{Conclusion}

Our study investigates the usage of various SSL speech encoders for a Spoken Language Understanding task in challenging circumstances characterized by a scarcity of both SLU and ASR training data and the low-resource characteristics of the targeted Tunisian Arabic dialect. Our findings emphasize the efficacy of SSL pre-trained speech encoders in such conditions, with notable success observed when employing the w2v-BERT 2.0 model, with 600 millions parameters trained on 4.5M hours of speech from 143 languages.
Additionally, we highlight the noteworthy performance of data2vec 2.0, pre-trained on English monolingual data, particularly excelling in handling semantically complex utterances. These outcomes collectively provide valuable information for advancing SLU methodologies, especially in resource-limited linguistic contexts.
Last, SAMU-XLSR provides very competitive results thanks to semantic enrichment made by the teacher student approach and in future work we plan to train a SAMU-w2vBERT 2.0 model to take benefit of the joint SAMU and w2v-BERT 2.0 capabilities.

\section{Acknowledgements}
This work was performed using HPC resources from GENCI-IDRIS (grant AD011012108R1) and received funding from the EU H2020 SELMA (grant No 957017), ANR TRADEF project (ANR-22-ASGC-0003) and ESPERANTO research and innovation programme under the Marie Skłodowska-Curie (grant No 101007666). 
We would like to especially thank our colleagues Sameer Khurana and Antoine Laurent for sharing their SAMU-XSLR model with us.

\appendix




\bibliography{custom}

\begin{thebibliography}{35}
\providecommand{\natexlab}[1]{#1}

\bibitem[{Babu et~al.(2021)Babu, Wang, Tjandra, Lakhotia, Xu, Goyal, Singh, von Platen, Saraf, Pino et~al.}]{babu2021xls}
Arun Babu, Changhan Wang, Andros Tjandra, Kushal Lakhotia, Qiantong Xu, Naman Goyal, Kritika Singh, Patrick von Platen, Yatharth Saraf, Juan Pino, et~al. 2021.
\newblock Xls-r: Self-supervised cross-lingual speech representation learning at scale.
\newblock pages arXiv--2111.

\bibitem[{Baevski et~al.(2022{\natexlab{a}})Baevski, Babu, Hsu, and Auli}]{Baevski2022EfficientSL}
Alexei Baevski, Arun Babu, Wei-Ning Hsu, and Michael Auli. 2022{\natexlab{a}}.
\newblock \href {https://api.semanticscholar.org/CorpusID:254685875} {Efficient self-supervised learning with contextualized target representations for vision, speech and language}.
\newblock In \emph{International Conference on Machine Learning}.

\bibitem[{Baevski et~al.(2022{\natexlab{b}})Baevski, Hsu, Xu, Babu, Gu, and Auli}]{baevski2022data2vec}
Alexei Baevski, Wei-Ning Hsu, Qiantong Xu, Arun Babu, Jiatao Gu, and Michael Auli. 2022{\natexlab{b}}.
\newblock Data2vec: A general framework for self-supervised learning in speech, vision and language.
\newblock In \emph{International Conference on Machine Learning}, pages 1298--1312. PMLR.

\bibitem[{Baevski et~al.(2020)Baevski, Zhou, Mohamed, and Auli}]{baevski2020wav2vec}
Alexei Baevski, Yuhao Zhou, Abdelrahman Mohamed, and Michael Auli. 2020.
\newblock wav2vec 2.0: A framework for self-supervised learning of speech representations.
\newblock \emph{Advances in neural information processing systems}.

\bibitem[{Barrault et~al.(2023)Barrault, Chung, Meglioli, Dale, Dong, Duppenthaler, Duquenne, Ellis, Elsahar, Haaheim et~al.}]{barrault2023seamless}
Lo{\"\i}c Barrault, Yu-An Chung, Mariano~Coria Meglioli, David Dale, Ning Dong, Mark Duppenthaler, Paul-Ambroise Duquenne, Brian Ellis, Hady Elsahar, Justin Haaheim, et~al. 2023.
\newblock Seamless: Multilingual expressive and streaming speech translation.
\newblock \emph{arXiv preprint arXiv:2312.05187}.

\bibitem[{Chen et~al.(2022)Chen, Wang, Chen, Wu et~al.}]{chen2022wavlm}
Sanyuan Chen, Chengyi Wang, Zhengyang Chen, Yu~Wu, et~al. 2022.
\newblock Wavlm: Large-scale self-supervised pre-training for full stack speech processing.
\newblock \emph{IEEE Journal of Selected Topics in Signal Processing}.

\bibitem[{Chung et~al.(2019)Chung, Hsu, Tang, and Glass}]{chung2019unsupervised}
Yu-An Chung, Wei-Ning Hsu, Hao Tang, and James Glass. 2019.
\newblock \href {https://doi.org/10.21437/Interspeech.2019-1473} {{An Unsupervised Autoregressive Model for Speech Representation Learning}}.
\newblock In \emph{Proc. Interspeech 2019}.

\bibitem[{Chung et~al.(2021)Chung, Zhang, Han, Chiu, Qin, Pang, and Wu}]{chung2021w2v}
Yu-An Chung, Yu~Zhang, Wei Han, Chung-Cheng Chiu, James Qin, Ruoming Pang, and Yonghui Wu. 2021.
\newblock W2v-bert: Combining contrastive learning and masked language modeling for self-supervised speech pre-training.
\newblock In \emph{ASRU 2021}.

\bibitem[{Conneau et~al.(2022)Conneau, Bapna, Zhang, Ma, von Platen, Lozhkov, Cherry, Jia, Rivera, Kale et~al.}]{conneau2022xtreme}
Alexis Conneau, Ankur Bapna, Yu~Zhang, Min Ma, Patrick von Platen, Anton Lozhkov, Colin Cherry, Ye~Jia, Clara Rivera, Mihir Kale, et~al. 2022.
\newblock \href {https://doi.org/10.21437/Interspeech.2022-10007} {{XTREME-S: Evaluating Cross-lingual Speech Representations}}.
\newblock In \emph{Proc. Interspeech 2022}, pages 3248--3252.

\bibitem[{Duquenne et~al.(2023)Duquenne, Schwenk, and Sagot}]{duquenne2023sentence}
Paul-Ambroise Duquenne, Holger Schwenk, and Beno{\^\i}t Sagot. 2023.
\newblock Sentence-level multimodal and language-agnostic representations.
\newblock \emph{arXiv preprint arXiv:2308.11466}.

\bibitem[{Evain et~al.(2021)Evain, Nguyen, Le, Boito, Mdhaffar, Alisamir, Tong et~al.}]{evain2021task}
Sol{\`e}ne Evain, Manh~Ha Nguyen, Hang Le, Marcely~Zanon Boito, Salima Mdhaffar, Sina Alisamir, Ziyi Tong, et~al. 2021.
\newblock Task agnostic and task specific self-supervised learning from speech with lebenchmark.
\newblock In \emph{NeurIPS}.

\bibitem[{Feng et~al.(2022)Feng, Yang, Cer, Arivazhagan, and Wang}]{feng2022language}
Fangxiaoyu Feng, Yinfei Yang, Daniel Cer, Naveen Arivazhagan, and Wei Wang. 2022.
\newblock Language-agnostic bert sentence embedding.
\newblock In \emph{Proceedings of the 60th Annual Meeting of the Association for Computational Linguistics (Volume 1: Long Papers)}, pages 878--891.

\bibitem[{Ghannay et~al.(2018)Ghannay, Caubri{\`{e}}re, Est{\`{e}}ve, Laurent, and Morin}]{Ghannay2018}
S.~Ghannay, A.~Caubri{\`{e}}re, Y.~Est{\`{e}}ve, A.~Laurent, and E.~Morin. 2018.
\newblock End-to-end named entity extraction from speech.
\newblock \emph{CoRR}, abs/1805.12045.

\bibitem[{Heffernan et~al.(2022)Heffernan, {\c{C}}elebi, and Schwenk}]{heffernan2022bitext}
Kevin Heffernan, Onur {\c{C}}elebi, and Holger Schwenk. 2022.
\newblock Bitext mining using distilled sentence representations for low-resource languages.
\newblock \emph{arXiv preprint arXiv:2205.12654}.

\bibitem[{Hsu et~al.(2021)Hsu, Bolte, Tsai, Lakhotia, Salakhutdinov, and Mohamed}]{hsu2021hubert}
Wei-Ning Hsu, Benjamin Bolte, Yao-Hung~Hubert Tsai, Kushal Lakhotia, Ruslan Salakhutdinov, and Abdelrahman Mohamed. 2021.
\newblock Hubert: Self-supervised speech representation learning by masked prediction of hidden units.
\newblock \emph{IEEE/ACM TASLP}.

\bibitem[{Khurana et~al.(2022)Khurana, Laurent, and Glass}]{khurana2022samu}
Sameer Khurana, Antoine Laurent, and James Glass. 2022.
\newblock Samu-xlsr: Semantically-aligned multimodal utterance-level cross-lingual speech representation.
\newblock \emph{Journal of Selected Topics in Signal Processing}.

\bibitem[{Laperri{\`e}re et~al.(2022)Laperri{\`e}re, Pelloin, Caubri{\`e}re, Mdhaffar, Camelin, Ghannay, Jabaian, and Est{\`e}ve}]{laperriere2022spoken}
Ga{\"e}lle Laperri{\`e}re, Valentin Pelloin, Antoine Caubri{\`e}re, Salima Mdhaffar, Nathalie Camelin, Sahar Ghannay, Bassam Jabaian, and Yannick Est{\`e}ve. 2022.
\newblock The spoken language understanding media benchmark dataset in the era of deep learning: data updates, training and evaluation tools.
\newblock In \emph{LREC}, pages 1595--1602.

\bibitem[{Liu et~al.(2021)Liu, Li, and Lee}]{liu2021tera}
Andy~T Liu, Shang-Wen Li, and Hung-yi Lee. 2021.
\newblock Tera: Self-supervised learning of transformer encoder representation for speech.
\newblock \emph{IEEE/ACM TASLP}.

\bibitem[{Masmoudi et~al.(2014)Masmoudi, Est{\`e}ve, Khmekhem, Bougares, and Belguith}]{masmoudi2014phonetic}
Abir Masmoudi, Yannick Est{\`e}ve, Mariem~Ellouze Khmekhem, Fethi Bougares, and Lamia~Hadrich Belguith. 2014.
\newblock Phonetic tool for the tunisian arabic.
\newblock In \emph{Spoken Language Technologies for Under-Resourced Languages}. Citeseer.

\bibitem[{Mdhaffar et~al.(2024)Mdhaffar, Bougars, De~mori, Zaiem, Ravanelli, and Estève}]{mdhaffar2024taric-slu}
Salima Mdhaffar, Fethi Bougars, Renato De~mori, Salah Zaiem, Mirco Ravanelli, and Yannick Estève. 2024.
\newblock Taric-slu: A tunisian dataset for spoken language understanding.
\newblock \emph{LREC}.

\bibitem[{Mdhaffar et~al.(2022)Mdhaffar, Pelloin, Caubri{\`e}re, Laperri{\`e}re, Ghannay, Jabaian, Camelin, and Est{\`e}ve}]{mdhaffar2022impact}
Salima Mdhaffar, Valentin Pelloin, Antoine Caubri{\`e}re, Ga{\"e}lle Laperri{\`e}re, Sahar Ghannay, Bassam Jabaian, Nathalie Camelin, and Yannick Est{\`e}ve. 2022.
\newblock Impact analysis of the use of speech and language models pretrained by self-supersivion for spoken language understanding.
\newblock In \emph{LREC 2022}.

\bibitem[{Ott et~al.(2019)Ott, Edunov, Baevski, Fan, Gross, Ng, Grangier, and Auli}]{ott2019fairseq}
Myle Ott, Sergey Edunov, Alexei Baevski, Angela Fan, Sam Gross, Nathan Ng, David Grangier, and Michael Auli. 2019.
\newblock fairseq: A fast, extensible toolkit for sequence modeling.
\newblock \emph{arXiv preprint arXiv:1904.01038}.

\bibitem[{Pratap et~al.(2023)Pratap, Tjandra, Shi, Tomasello, Babu, Kundu, Elkahky, Ni, Vyas, Fazel-Zarandi et~al.}]{pratap2023scaling}
Vineel Pratap, Andros Tjandra, Bowen Shi, Paden Tomasello, Arun Babu, Sayani Kundu, Ali Elkahky, Zhaoheng Ni, Apoorv Vyas, Maryam Fazel-Zarandi, et~al. 2023.
\newblock Scaling speech technology to 1,000+ languages.
\newblock \emph{arXiv preprint arXiv:2305.13516}.

\bibitem[{Radford et~al.(2023)Radford, Kim, Xu, Brockman, McLeavey, and Sutskever}]{radford2023robust}
Alec Radford, Jong~Wook Kim, Tao Xu, Greg Brockman, Christine McLeavey, and Ilya Sutskever. 2023.
\newblock Robust speech recognition via large-scale weak supervision.
\newblock In \emph{International Conference on Machine Learning}, pages 28492--28518. PMLR.

\bibitem[{Ravanelli et~al.(2021)Ravanelli, Parcollet, Plantinga, Rouhe et~al.}]{ravanelli2021speechbrain}
Mirco Ravanelli, Titouan Parcollet, Peter Plantinga, Aku Rouhe, et~al. 2021.
\newblock Speechbrain: A general-purpose speech toolkit.
\newblock \emph{arXiv preprint arXiv:2106.04624}.

\bibitem[{Schneider et~al.(2019)Schneider, Baevski, Collobert, and Auli}]{schneider2019wav2vec}
Steffen Schneider, Alexei Baevski, Ronan Collobert, and Michael Auli. 2019.
\newblock wav2vec: Unsupervised pre-training for speech recognition.
\newblock \emph{Interspeech 2019}.

\bibitem[{Searle(1969)}]{searle_1969}
John~R. Searle. 1969.
\newblock \href {https://doi.org/10.1017/CBO9781139173438} {\emph{Speech Acts: An Essay in the Philosophy of Language}}.
\newblock Cambridge University Press.

\bibitem[{Shi et~al.(2023)Shi, Berrebbi, Chen, Chung, Hu, Huang, Chang, Li, Mohamed, Lee et~al.}]{shi2023ml}
Jiatong Shi, Dan Berrebbi, William Chen, Ho-Lam Chung, En-Pei Hu, Wei~Ping Huang, Xuankai Chang, Shang-Wen Li, Abdelrahman Mohamed, Hung-yi Lee, et~al. 2023.
\newblock Ml-superb: Multilingual speech universal performance benchmark.
\newblock \emph{arXiv}.

\bibitem[{Toyin et~al.(2023)Toyin, Djanibekov, Kulkarni, and Aldarmaki}]{toyin2023artst}
Hawau~Olamide Toyin, Amirbek Djanibekov, Ajinkya Kulkarni, and Hanan Aldarmaki. 2023.
\newblock Artst: Arabic text and speech transformer.
\newblock \emph{arXiv preprint arXiv:2310.16621}.

\bibitem[{Tur and De~Mori(2011)}]{tur2011spoken}
Gokhan Tur and Renato De~Mori. 2011.
\newblock \emph{Spoken language understanding: Systems for extracting semantic information from speech}.
\newblock John Wiley \& Sons.

\bibitem[{Wang et~al.(2021)Wang, Rivi{\`e}re, Lee, Wu, Talnikar, Haziza, Williamson, Pino, and Dupoux}]{wang2021voxpopuli}
Changhan Wang, Morgane Rivi{\`e}re, Ann Lee, Anne Wu, Chaitanya Talnikar, Daniel Haziza, Mary Williamson, Juan Pino, and Emmanuel Dupoux. 2021.
\newblock Voxpopuli: A large-scale multilingual speech corpus for representation learning, semi-supervised learning and interpretation.
\newblock In \emph{ACL}.

\bibitem[{Wang et~al.(2023)Wang, Li, Guo, Qiao, Li, Shang, Wei, Tao, Zhang, and Yang}]{wang2023whislu}
Minghan Wang, Yinglu Li, Jiaxin Guo, Xiaosong Qiao, Zongyao Li, Hengchao Shang, Daimeng Wei, Shimin Tao, Min Zhang, and Hao Yang. 2023.
\newblock Whislu: End-to-end spoken language understanding with whisper.
\newblock In \emph{Proc. Interspeech}, volume 2023, pages 770--774.

\bibitem[{Wang et~al.(2005)Wang, Deng, and Acero}]{wang2005spoken}
Ye-Yi Wang, Li~Deng, and Alex Acero. 2005.
\newblock Spoken language understanding.
\newblock \emph{Signal Processing Magazine}.

\bibitem[{Yadav et~al.(2020)Yadav, Ghosh, Yu, and Shah}]{yadav2020end}
Hemant Yadav, Sreyan Ghosh, Yi~Yu, and Rajiv~Ratn Shah. 2020.
\newblock End-to-end named entity recognition from english speech.
\newblock \emph{Interspeech}.

\bibitem[{Yang et~al.(2021)Yang, Chi, Chuang, Lai, Lakhotia, Lin, Liu, Shi, Chang, Lin et~al.}]{yang2021superb}
Shu~Wen Yang, Po~Han Chi, Yung~Sung Chuang, Cheng I~Jeff Lai, Kushal Lakhotia, Yist~Y Lin, Andy~T Liu, Jiatong Shi, Xuankai Chang, Guan~Ting Lin, et~al. 2021.
\newblock Superb: Speech processing universal performance benchmark.
\newblock In \emph{22nd Annual Conference of the International Speech Communication Association, INTERSPEECH 2021}, pages 3161--3165. International Speech Communication Association.

\end{thebibliography}

\section{Appendix: URLs of all models used in this study}
\label{sec:appendix}
For reproducibility of all results obtained in this paper, we put in the table \ref{tab:links} the url for each model.

\begin{table*}[]
\centering
\caption{URLs of all models used in this study}
\label{tab:links}
\begin{tabular}{|l|l|}
\hline
Speech encoder          & URL                                                                    \\ \hline \hline
LeBenchmark-7K & \href{https://huggingface.co/LeBenchmark/wav2vec2-FR-7K-large}{https://huggingface.co/LeBenchmark/wav2vec2-FR-7K-large}                \\ \hline
English lv60   & \href{https://huggingface.co/facebook/wav2vec2-large-lv60}{https://huggingface.co/facebook/wav2vec2-large-lv60}                     \\ \hline
HuBERT         & \href{https://huggingface.co/facebook/hubert-large-ll60k}{https://huggingface.co/facebook/hubert-large-ll60k}                      \\ \hline
WavLM          & \href{https://huggingface.co/microsoft/wavlm-large}{https://huggingface.co/microsoft/wavlm-large}                             \\ \hline
Data2vec 2.0   & \href{https://github.com/facebookresearch/fairseq/tree/main/examples/data2vec}{https://github.com/facebookresearch/fairseq/tree/main/examples/data2vec} \\ \hline
VP-100K        & \url{https://huggingface.co/facebook/wav2vec2-large-100k-voxpopuli}           \\ \hline
XLS-R-128      & \href{https://huggingface.co/facebook/wav2vec2-xls-r-300m}{https://huggingface.co/facebook/wav2vec2-xls-r-300m}                     \\ \hline
MMS            & \href{https://huggingface.co/facebook/mms-300m}{https://huggingface.co/facebook/mms-300m}                                \\ \hline
MMS-1B         & \href{https://huggingface.co/facebook/mms-1b}{https://huggingface.co/facebook/mms-1b}                                  \\ \hline
w2v-BERT 2.0   & \href{https://huggingface.co/facebook/w2v-bert-2.0}{https://huggingface.co/facebook/w2v-bert-2.0}                            \\ \hline
SAMU-XLSR      & \small SAMU-XLSR is not yet publicly available; it has been shared by \citet{khurana2022samu}                                                                    \\ \hline
SONAR-ENG      & \href{https://github.com/facebookresearch/SONAR}
{https://github.com/facebookresearch/SONAR}\\ \hline
SONAR-FRA      & \href{https://github.com/facebookresearch/SONAR}
{https://github.com/facebookresearch/SONAR}\\ \hline
SONAR-ARB      & \href{https://github.com/facebookresearch/SONAR}    {https://github.com/facebookresearch/SONAR}                           \\ \hline \hline
Whisper-small      &  \href{https://huggingface.co/openai/whisper-small}
{https://huggingface.co/openai/whisper-small} \\ \hline
Whisper-medium      & \href{https://huggingface.co/openai/whisper-medium}
{https://huggingface.co/openai/whisper-medium}\\ \hline 
\end{tabular}
\end{table*}

\end{document}